\documentclass[runningheads]{llncs}
\usepackage{graphicx, framed}
\usepackage{color}
\usepackage{amsmath, amsfonts}
\usepackage{mathtools}
\usepackage{url}
\usepackage{algpseudocode}
\usepackage{algorithm2e}

\DeclarePairedDelimiterX{\infdivx}[2]{(}{)}{%
  #1\;\delimsize\|\;#2%
}
\newcommand{\infdiv}{\mathbf{D}_{KL}\infdivx}

\newcommand\ie{\emph{i.e.}\ }
\newcommand\eg{\emph{e.g.}\ }

\newcommand\dskip[1]{}

\newcommand\flip[1]{\overline{#1}}
\newcommand\Flip{\mathrm{Flip}}
\newcommand\Flop{\mathrm{Flop}}

\newcommand\softmax{\mathrm{Softmax}}
\newcommand\uscore[2]{\mathbf{#1}_{\text{#2}}}

\newcommand\myvrow{\text{---}}
\newcommand\SATG{\mathcal{G}}

\usepackage{tikz}
\usetikzlibrary{shapes,positioning,calc,arrows.meta}
\tikzstyle{lit}=[]
\tikzstyle{clause}=[]
\tikzstyle{lc}=[]
\tikzstyle{ll}=[dashed]
\tikzstyle{occ}=[-]
\tikzstyle{flip}=[dashed]

\tikzstyle{send}=[-{Latex[length=1mm, width=2.2mm]}]
\tikzstyle{sendll}=[densely dashed, -{Latex[length=0.72mm, width=1.6mm]}, bend right=10]
\tikzstyle{pcloud}=[cloud, draw=black, aspect=1]
\tikzstyle{mcloud}=[cloud, draw=black, aspect=2.5]
\tikzstyle{boxed}=[ellipse, draw=black]
\tikzstyle{bhead}=[-{Latex[length=1mm, width=2.2mm]}]

\begin{document}
\title{Guiding High-Performance SAT Solvers with Unsat-Core Predictions}
%
%
\author{Daniel Selsam\inst{1}%
\thanks{This paper describes work performed while the first author was at Microsoft Research.}
\and Nikolaj Bj{\o}rner\inst{2}}

\authorrunning{D. Selsam and N. Bj{\o}rner}
%
\institute{Stanford University, Stanford CA 94305, USA \and Microsoft Research, Redmond WA 98052, USA}
%
\maketitle              

\begin{abstract}
  The \emph{NeuroSAT} neural network architecture was introduced
  in~\cite{selsam2018learning} for predicting properties of
  propositional formulae. When trained to predict the satisfiability
  of toy problems, it was shown to find solutions and unsatisfiable
  cores on its own. However, the authors saw ``no obvious path'' to
  using the architecture to improve the state-of-the-art.  In this
  work, we train a simplified NeuroSAT architecture to directly
  predict the unsatisfiable cores of real problems. We modify several
  high-performance SAT solvers to periodically replace their variable
  activity scores with NeuroSAT's prediction of how likely the
  variables are to appear in an unsatisfiable core. The modified
  MiniSat solves 10\% more problems on SATCOMP-2018 within the
  standard 5,000 second timeout than the original does. The modified
  Glucose solves 11\% more problems than the original, while the
  modified Z3 solves 6\% more. The gains are even greater when the training is specialized for
  a specific distribution of problems; on a benchmark of hard problems
  from a scheduling domain, the modified Glucose solves 20\% more
  problems than the original does within a one-hour timeout.
  Our results demonstrate that NeuroSAT
  can provide effective guidance to
  high-performance SAT solvers on real problems.
\end{abstract}

\section{Introduction}\label{sec:intro}

Over the past decade, neural networks have dramatically advanced the
state of the art on many important problems, most notably object
recognition~\cite{krizhevsky2012imagenet}, speech
recognition~\cite{hinton2012deep}, and machine
translation~\cite{wu2016google}.  There have also been several
attempts to apply neural networks to problems in discrete search, such
as program synthesis~\cite{parisotto2016neuro,devlin2017robustfill},
first-order theorem proving~\cite{irving2016deepmath,loos2017deep} and
higher-order theorem
proving~\cite{whalen2016holophrasm,wang2017premise,kaliszyk2017holstep,huang2018gamepad}.
More recently, \cite{selsam2018learning}
introduce
a neural network architecture designed for satisfiability problems,
called \emph{NeuroSAT}, and show that when trained to predict
satisfiability on toy problems, it learns to find solutions and unsatisfiable
cores on its own. Moreover, the neural network is iterative, and the
authors show that by running for many more iterations at test time, it
can solve problems that are bigger and even from completely different
domains than the problems it was trained on. While these results may
be intriguing, the authors' motivation was to study the capabilities
of neural networks rather than to solve real SAT problems, and they
admit to seeing ``no obvious path'' to beating existing SAT solvers.

In this work, we make use of the NeuroSAT architecture, but whereas it
was originally used as an end-to-end solver on toy problems, here we
use it to help inform variable branching decisions within
high-performance SAT solvers on real problems. Given this goal, the
main design decision becomes how to produce data to train the
network. Our approach is to generate a supervised dataset mapping
unsatisfiable problems to the variables in their unsatisfiable cores.
Note that perfect predictions would not always
yield a useful variable branching heuristic; for some problems,
the smallest core may include every
variable, and of course for satisfiable problems, there
are no cores at all. Thus, our approach is pragmatic; we rely on
NeuroSAT predicting \emph{imperfectly}, and hope that the probability
NeuroSAT assigns to a given variable being in a core correlates well
with that variable being good to branch on.

The next biggest design decision is how to make use of the predictions
inside a SAT solver. Even if we wanted to query NeuroSAT for every
variable branching decision, doing so would have severe performance
implications, particularly for large problems.  A SAT solver makes
tens of thousands of assignments every second, whereas even with an
on-device GPU, querying NeuroSAT on an industrial-sized problem may
take hundreds or even thousands of milliseconds.  We settle for
complementing---rather than trying to replace---the efficient variable
branching heuristics used by existing solvers. All three solvers we
extend (MiniSat, Glucose, Z3) use the Exponential Variable
State-Independent Decaying Sum (EVSIDS) heuristic, which involves
maintaining activity scores for every variable and branching on the
free variable with the highest score. The only change we make is that
we periodically query NeuroSAT on the \emph{entire} problem (i.e. not
conditioning on the current trail), and set all variable activity
scores at once in proportion to how likely NeuroSAT thinks the
variable is to be involved in an unsat core. We refer to our
integration strategy as \emph{periodic refocusing}.  We remark that
the base heuristics are already strong, and they may only need an
occasional, globally-informed reprioritization to yield substantial
improvements.

We summarize our pipeline:

\begin{enumerate}
\item Generate many unsatisfiable problems by decimating existing problems.
\item For each such problem, generate a DRAT proof, and extract the variables that appear in the unsat core.
\item Train NeuroSAT (henceforth NeuroCore) to map unsatisfiable problems to the variables in the core.
\item Instrument state-of-the-art solvers (MiniSat, Glucose, Z3) to query NeuroCore periodically (using the
  original and the learnt clauses), and to reset their variable activity
  scores according to NeuroCore's predictions.
\end{enumerate}

As a result of these modifications, the MiniSat solver solves 10\%
more problems on SATCOMP-2018 within the standard 5,000 second
timeout. The modified Glucose 4.1 solves 11\% more problems than the
original, while the modified Z3 solves 6\% more.  The gains are even
greater when the training is specialized for a specific distribution
of problems; our training set included (easy) subproblems of a
collection of hard scheduling problems, and on that collection of hard
problems the modified Glucose solves 20\% more problems than the
original does within a one-hour timeout.  Our results demonstrate that
NeuroSAT (and in particular, NeuroCore) can provide effective guidance
to high-performance SAT solvers on real problems.  All scripts and
sources associated with NeuroCore are available
from~\url{https://github.com/dselsam/neurocore-public}.

\section{Data generation}\label{sec:data}

As discussed in \S\ref{sec:intro}, we want to train our neural network
architecture to predict which variables will be involved in unsat
cores. Unfortunately, there are only roughly one thousand unsatisfiable
problems across all SATCOMP competitions, and a network trained on
such few examples would be unlikely to generalize well to unseen
problems. We overcome this limitation and generate a dataset
containing over 150,000 different problems with labeled cores by
considering unsatisfiable \emph{subproblems} of existing problems.

Specifically, we generate training data as follows. We use the
distributed execution framework ray~\cite{moritz2018ray} to coordinate
one driver and hundreds of workers distributed over several
machines. The driver maintains a queue of (sub)problems, and begins by
enqueuing all problems from SATCOMP (through 2017 only) as well as a
few hundred hard scheduling problems. It might help to initialize with even
more problems, but we did not find it necessary to do so. Whenever a
worker becomes free, the driver dequeues a problem and passes it to
the worker. The worker tries to solve it using Z3 with a fixed timeout
(we used 60 seconds). If Z3 returns \emph{sat}, it does nothing, but
if Z3 returns \emph{unsat}, it passes the generated DRAT
proof~\cite{wetzler2014drat} to DRAT-trim~\cite{wetzler2014drat} to
determine which of the original clauses were used in the proof. It
then computes the variables in the core by traversing the clauses in
the core, and finally generates a single datapoint in a format suitable for
the neural network architecture we will describe in \S\ref{sec:architecture}.
If Z3 returns \emph{unknown}, the worker uses a relatively expensive, hand-engineered variable branching heuristic (specifically, Z3's
implementation of the March heuristic~\cite{Mijnders2010SymbiosisOS}) and returns the two
subproblems to the driver to be added to the queue.

This process generates one datapoint roughly every 60 seconds per
worker.  Some of the original problems are very difficult, and so the
process may not terminate in a reasonable amount of time; thus we
stopped it once we had generated 150,000 datapoints.


Note that our data generation process is not guaranteed to generate
diverse cores. To the extent that March is successful in selecting
variables to branch on that are in the core, the cores of the two
subproblems will be different; if it fails to do this, then the cores
of the two subproblems may be the same (though the non-core clauses
will still be different). We remark that there are many other ways one
might augment the dataset, for example by including additional
problems from synthetic distributions, or by directly perturbing the
signs of the literals in the existing problems. However, our simple
approach proved sufficient.

We stress that predicting the (binary) presence of variables in the
core is simplistic. As mentioned in \S\ref{sec:intro}, for some
problems, the smallest core may include every variable, in which case
the datapoint for that problem would contain no information. Even if
only a small fraction of variables are in the core, it may still be
that only a small fraction of those core variable would make good
branches. A more sophisticated approach would analyze the full DRAT
proof and calculate a more nuanced score for each variable that
reflects its importance in the proof. However, as we will see in
\S\ref{sec:experiments}, our simplistic approach of predicting the
variables in the core proved sufficient to achieve compelling results.

\section{Neural Network Architecture}\label{sec:architecture}

\paragraph{Background on neural networks.}
Before describing our simplified version of the NeuroSAT architecture,
we provide elementary background on neural networks. A neural network
can be thought of as a computer program that is differentiable with
respect to a set of real-valued, unknown parameters.  There may be
thousands, millions, or even billions of such parameters, and it would
be impossible to specify them by hand. Instead, the practitioner
specifies a second differentiable program called the \emph{loss
  function}, which takes a collection of input/output pairs (\ie
training data), runs the neural network on the inputs, and computes a
scalar score that measures how much the neural network's outputs
disagree with the true outputs. Numerical optimization is then used
to find values of the unknown parameters that make the loss function
as small as possible.

The basic building block of neural networks is the
multilayer perceptron (MLP), also called a feed-forward network or a
fully-connected network.  An MLP takes as input a vector $x \in
\mathbb{R}^{d_{in}}$ for a fixed $d_{in}$, and outputs a vector $y \in
\mathbb{R}^{d_{out}}$ for a fixed $d_{out}$.  It computes $y$ from $x$
by applying a sequence of (parameterized) affine transformations, each
but the last followed by a component-wise nonlinear function called an
\emph{activation function}.  The most common activation function
(which we use in this work) is the rectified linear unit (ReLU), which
is the identity function on positive numbers and sets all negative
numbers to zero.

\paragraph{Notation.} We use function-call notation to denote the application of MLPs,
where the different arguments to the MLP are implicitly concatenated. Thus if
$M : \mathbb{R}^{d_1 + d_2} \to \mathbb{R}^{d_{out}}$ is an MLP
and $x_1 \in \mathbb{R}^{d_1}, x_2 \in \mathbb{R}^{d_2}$ are vectors,
we write $M(x_1, x_2) \in \mathbb{R}^{d_{out}}$
to denote the result of applying the MLP $M$ to the concatenation of $x_1$ and $x_2$.
For performance reasons, one almost never applies an MLP to an individual vector, and instead
applies it to a \emph{batch} of vectors of the same dimension concatenated into a matrix.
Thus if $X_1 \in \mathbb{R}^{k \times d_1}, X_2 \in \mathbb{R}^{k \times d_2}$,
we write $M(X_1, X_2) \in \mathbb{R}^{k \times d_{out}}$ to denote the result of first concatenating
$X_1$ and $X_2$ into a $\mathbb{R}^{k \times (d_1 + d_2)}$ matrix,
applying $M$ to the each of the $k$ rows separately and then concatenating the $k$ results back into a matrix.

\paragraph{NeuroCore architecture.}
We now describe our simplified version of the NeuroSAT
architecture.
We represent a Boolean formula in CNF with $n_v$ variables and $n_c$ clauses by
an $n_c \times 2n_v$ sparse matrix $\SATG$,
where the $(i, j)$th element is 1 if and only if the $i$th clause contains the $j$th literal.
For example, we represent the formula \( \underbrace{(x_1 \vee x_2 \vee x_3)}_{c_1} \wedge \underbrace{(x_1 \vee \flip{x_2} \vee \flip{x_3})}_{c_2} \)
as the following $2 \times 6$ (sparse) matrix:

\[
\begin{array}{rcl}
\SATG & :\ \ \ \ \   &
\begin{array}{l|llllll}
    & x_1 & x_2  & x_3 & \flip{x_1} & \flip{x_2} & \flip{x_3} \\
\hline
  c_1 & 1   & 1    & 1   &  0        &  0         &  0         \\
  c_2 & 1   & 0    & 0   &  0        &  1         &  1
\end{array}
\end{array}
\]
Our neural network itself is made up of three standard MLPs:
\[
  \uscore{C}{update} :  \mathbb{R}^{2d} \to \mathbb{R}^d,
  \uscore{L}{update} :  \mathbb{R}^{3d} \to \mathbb{R}^d,
  \uscore{V}{proj}   :  \mathbb{R}^{2d} \to \mathbb{R}
\]
where $d$ is a fixed hyperparameter (we used $d=80$).
The network computes forward as follows. First, it initializes two
matrices $C \in \mathbb{R}^{n_c \times d}$ and $L \in \mathbb{R}^{2n_v
  \times d}$ to all ones. Each row of $C$ corresponds to a clause,
while each row of $L$ corresponds to a literal:

\[
C =
\begin{bmatrix}
  \myvrow & c_1 & \myvrow  \\
          &  \vdots &                      \\
    \myvrow & c_{n_c} & \myvrow
\end{bmatrix}
\in \mathbb{R}^{n_c \times d}, \qquad
L =
\begin{bmatrix}
  \myvrow & x_1 & \myvrow  \\
  &        \vdots &       \\
  \myvrow & x_{n_v} & \myvrow  \\
  \myvrow & \flip{x_1} & \myvrow  \\
  & \vdots& \\
  \myvrow & \flip{x_{n_v}} & \myvrow
\end{bmatrix} \in \mathbb{R}^{2n_v \times d}
\]
We refer to the row corresponding to a clause $c$ or a literal $\ell$
as the \emph{embedding} of that clause or literal.
Note that for notational convenience,
we conflate clauses and literals with their embeddings,
so \eg the symbol $c$ may refer to the actual clause or to the row of $C$ that embeds the clause.

Define
the operation $\Flip$ to swap the first half of the rows of a matrix with the
second half, so that in $\Flip(L)$, each literal's row is swapped with its negation's:
\[
\Flip(L) =
\begin{bmatrix}
  \myvrow & \flip{x_1} & \myvrow  \\
  & \vdots& \\
  \myvrow & \flip{x_{n_v}} & \myvrow \\
  \myvrow & x_1 & \myvrow  \\
  &        \vdots &       \\
  \myvrow & x_{n_v} & \myvrow
\end{bmatrix}
\in \mathbb{R}^{2n_v \times d}
\]

After initializing $C$ and $L$, the network performs $T$ iterations of ``message passing'' (we used $T=4$),
where a single iteration consists of two updates.
First, each clause updates its embedding based on the current embeddings of the literals it contains:
\( \forall c, c \gets \uscore{C}{update}\left( c, \sum_{\ell \in c} \ell \right) \).
Next, each literal updates its embedding based on the current embeddings of the clauses it occurs in, as well
as the current embedding of its negation: \(
\forall \ell, \ell \gets \uscore{L}{update}( \ell, \sum_{\ell \in c} c, {\flip{\ell}} ) \).
We can express these updates compactly and implement them efficiently using the matrix $\SATG$ and the $\Flip$ operator:
\begin{align*}
  C & \gets \uscore{C}{update}\left( C, \SATG L \right)  \\
  L & \gets \uscore{L}{update}\left( L, \SATG^{\top}C, \Flip(L) \right)
\end{align*}
Define
the operation $\Flop$ to concatenate the first half of the rows of a matrix with the
second half along the second axis,
so that in $\Flop(L)$, the two vectors corresponding to the same variable are concatenated:
\[
\Flop(L) =
\begin{bmatrix}
\myvrow & x_1    & \myvrow & \hspace{0.2cm} & \myvrow & \flip{x_1} & \myvrow  \\
&  &  & \vdots &  &  &   \\
\myvrow & x_{n_v} & \myvrow & \hspace{0.2cm} & \myvrow & \flip{x_{n_v}} & \myvrow
\end{bmatrix}
\in \mathbb{R}^{n_v \times 2d}
\]
After $T$ iterations, the network flops $L$ to produce the matrix $V \in \mathbb{R}^{n_v \times 2d}$,
and then projects $V$ into an $n_v$-dimensional vector $\hat{v}$ using the third MLP, $\uscore{V}{proj}$:
\[
  \hat{v} \gets \uscore{V}{proj}(V) \in \mathbb{R}^{n_v}
\]

\begin{figure}

  \centering
\parbox{2.2in}{
   \vspace{-115pt}
  \begin{framed}
$
\begin{array}{l}
  \text{Initialize: }  \\
  \qquad C \gets \mathbf{1} \in \mathbb{R}^{n_c \times d} \\
  \qquad L \gets \mathbf{1} \in \mathbb{R}^{2n_v \times d}  \\
  \text{$T$ times: }  \\
  \qquad C \gets \uscore{C}{update}\left( C, \SATG L \right) \\
  \qquad L \gets \uscore{L}{update}\left( L, \SATG^{\top}C, \Flip(L) \right) \\
  \text{Finally: } \\
  \qquad V \gets \Flop(L) \in \mathbb{R}^{n_v \times 2d} \\
  \qquad \hat{v} \,\, \gets \uscore{V}{proj}(V) \in \mathbb{R}^{n_v}
\end{array}
$
\end{framed}
}
\qquad
  \begin{tikzpicture}[node distance=1cm, auto, draw=black]
    \node[draw=black] (C) {$C$};
    \node[right=3.0cm of C, draw=black] (L) {$L$};

    \node[below=0.3cm of C] (GHOST0) {};
    \node[below=1.0cm of C] (GHOST1) {};
    \node[below=1.7cm of C] (GHOST2) {};

    \node[below=2.0cm of C, draw=red] (CU) {$\uscore{C}{update}$};
    \node[below=2.0cm of L, draw=red] (LU) {$\uscore{L}{update}$};

    \node[right=1.1cm of GHOST0, draw=black] (LC) {$\mathcal{G} L$};
    \node[right=1.0cm of GHOST2, draw=black] (CL) {$\mathcal{G}^{\top} C$};
    \node[right=2.0cm of GHOST1, draw=black] (FL) {$\Flip(L)$};

    \node[below=0.5cm of LU, draw=black] (V) {$\Flop(L)$};
    \node[left=0.5cm of V, draw=red] (VP) {$\uscore{V}{proj}$};
    \node[left=0.5cm of VP, draw=black] (VH) {$\hat{v}$};

    \draw[->] (L) to (LC);
    \draw[->] (C) to (CL);

    \draw[->] (L) to (LU);
    \draw[->] (CL) to (LU);
    \draw[->] (C) to (CU);
    \draw[->] (LC) to (CU);
    \draw[->] (L) to (FL);
    \draw[->] (FL) to (LU);

    \draw[->, bend left=40pt] (CU) to (C);
    \draw[->, bend right=40pt] (LU) to (L);

    \draw[->] (LU) to (V);
    \draw[->] (V) to (VP);
    \draw[->] (VP) to (VH);
\end{tikzpicture}
\caption{An overview of the NeuroCore architecture\label{fig:arch}}
\end{figure}
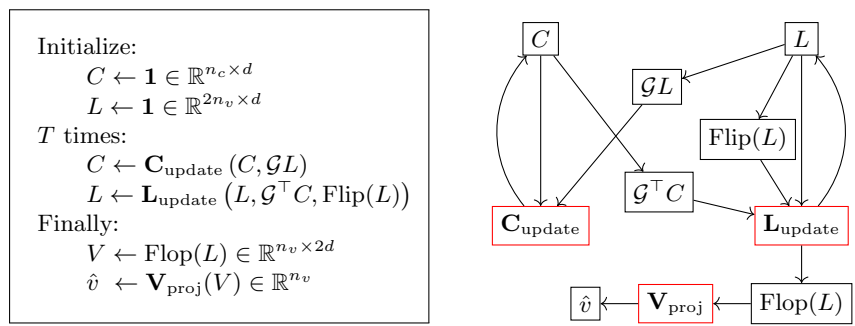

The vector $\hat{v}$ is the output of NeuroCore, and consists of
a numerical score for each variable, which can be passed to the softmax
function to define a probability distribution $\hat{p}$ over the variables.
During training, we turn each labeled bitmask over
variables into a probability distribution $p^*$ by assigning
uniform probability to each variable in the core and zero probability
to the others. We optimize the three MLPs all at once to
minimize the Kullback-Leibler divergence~\cite{kullback1951information}:
\[
  \infdiv{p^*}{\hat{p}}
=
  \sum_{i=1}^{n_v} p^*_i \log \left( p^*_i / \hat{p}_i \right)
\]
Figure~\ref{fig:arch} summarizes the architecture.

\paragraph{Comparison to the original NeuroSAT.}
While the original NeuroSAT architecture was designed to solve small
problems end-to-end, ours is designed to provide cheap, heuristic
guidance on (potentially) large problems. Accordingly, our network
differs from the original in a few key ways. First, ours only runs for
4 iterations at both train and test time, whereas the original was
trained with 26 iterations and ran for upwards of a thousand
iterations at test time. Second, our update networks are simple
MLPs, whereas the original used Long Short-Term
Memories (LSTMs)~\cite{hochreiter1997lstm}. Third, as discussed above,
ours is trained with supervision at every variable and
outputs a vector $\hat{v} \in
\mathbb{R}^{n_v}$, whereas the original is trained with only a single
bit of supervision and accordingly only outputs a single scalar.

\paragraph{Training NeuroCore.}
As we discussed in \S\ref{sec:intro}, our goal is not to learn a
perfect core predictor, but rather only to learn a coarse heuristic
that broadly assigns higher score to more important variables.  Thus,
fine-tuning the network is relatively unimportant, and we only ever
trained with a single set of hyperparameters. We used the
ADAM optimizer~\cite{kingma2014adam} with a constant learning rate
of $10^{-4}$, and trained asynchronously with 20 GPUs for under an hour,
using distributed TensorFlow~\cite{abadi2016tensorflow}.

\section{Hybrid Solving: Extending CDCL with NeuroCore}\label{sec:hybrid}

\paragraph{Background on CDCL.}

Modern SAT solvers are based on the Conflict-Driven Clause Learning
(CDCL) algorithm~\cite{marques1999grasp,knuth2015sat}. Before
explaining how we integrate NeuroCore with CDCL solvers , we briefly
summarize the parts of CDCL that are relevant to our work.  At a high
level, a CDCL solver works as follows. It maintains a \emph{trail} of
literals that have been given tentative assignments, and continues to
assign variables and propagate the implications until reaching a
contradiction. It then analyzes the cause of the contradiction and
learns a \emph{conflict clause} that is implied by the existing
clauses and that would have helped avoid the current
conflict. Finally, it pops variables off the trail until all but one
of the literals in the learnt clause have been set to false,
propagates the learnt clause, and continues from there. Most CDCL
solvers also periodically \emph{restart} (clear the
trail), and also periodically simplify the clauses in various ways.

There are many crucial, heuristic decisions that a CDCL must make,
such as which variable to branch on next, what polarity to set it to,
which learned clauses to prune and when, and also when to restart.  We
only consider the first decision in this work: which variable to branch on next.  This
decision has been the subject of
intense study for decades and many approaches have been proposed.
See~\cite{biere2015evaluating} for a comprehensive overview.  MiniSat, Glucose and Z3 all
implement variants of the Variable State-Independent Decaying Sum
(VSIDS) heuristic (first introduced in~\cite{moskewicz2001chaff}) called
Exponential-VSIDS (EVSIDS). The EVSIDS score of a variable $x$ after the $t$th conflict is defined
by:
\begin{align*}
  \mathbf{InConflict}(x, i) & =
  \begin{cases} 1 &\text{$x$ was involved in the $i$th conflict} \\
    0 & \text{otherwise} \end{cases} \\
  \mathbf{EVSIDS}(x, t) & = \sum_{i} \mathbf{InConflict}(x, i) \rho^{t-i}
\end{align*}
where $\rho < 1$ is a hyperparameter.
Intuitively, the EVSIDS score of a variable measures how many conflicts the variable has been
involved in, with more recent conflicts weighted much more than past
conflicts. As we will discuss in \S\ref{sec:hybrid}, our approach is
to periodically reset these EVSIDS scores based on the outputs of
NeuroCore.

\paragraph{Integrating NeuroCore.}
As discussed in \S\ref{sec:intro}, it is too expensive to query
NeuroCore for every variable branching decision, and so we settle for
querying periodically on the entire problem (\ie not conditioning on the trail)
and replacing the variable
activity scores with NeuroCore's prediction. We now describe this
process in detail.

When we query NeuroCore, we build the sparse clause-literal adjacency matrix
$\SATG$ (see \S\ref{sec:architecture}) as follows. First, we collect all
non-eliminated variables that are not units at level 0. These are the
only variables we tell NeuroCore about. Second, we collect all the
clauses that we plan to tell NeuroCore about. We would like to tell
NeuroCore about all the clauses, both original and learnt, but the
size of the problem can get extremely large as the solver accumulates
learnt clauses. At some point the problem would no longer fit in
GPU memory, and it might be undesirably expensive even before that point.
After collecting the original clauses, we traverse the learned clauses
in ascending size order, collecting clauses until the number of
literals plus the number of clauses plus the number of cells (\ie literal occurrences in clauses) exceed a
fixed cutoff (we used 10 million). If a problem is so big that the
original clauses already exceed this cutoff, then for simplicity we do
not query NeuroCore at all, although we could have still queried it on
random subsets of the clauses. Finally, we traverse the chosen clauses
to construct $\SATG$. Note that because of
the learned clauses, the eliminated variables, and the discovered units,
NeuroCore is shown a substantially different
graph on each query even though we do not condition on the trail.

NeuroCore then returns a vector $\hat{v} \in \mathbb{R}^{n_v}$, where
a higher score for a variable indicates that NeuroCore thinks the corresponding
variable is more likely to be in the core. We turn $\hat{v}$ into a probability
distribution by dividing it by a scalar temperature parameter $\tau$ (we used 0.25)
and taking the softmax, and then we scale the resulting vector by the number of
variables in the problem, and additionally by a fixed constant
$\kappa$ (we used $10^4$). Finally, we replace all the EVSIDS scores at once:\footnote{In MiniSat, this involves setting the activity vector to these values, resetting the variable increment to 1.0, and rebuilding the order-heap.}
\[
\forall i,
  \mathbf{EVSIDS}(x_i, t) \gets \softmax(\hat{v} / \tau)_i n_v \kappa
\]
Note that the decay factor $\rho$ is often rather small (MiniSat uses $\rho=0.95$),
and to a first approximation solvers average ten thousand conflicts per second, so
these scores decay to 0 in only a fraction of a second. However, such
an intervention can still have a powerful effect by refocusing EVSIDS
on a more important part of the search space.
We refer to our integration strategy as \emph{periodic refocusing} to stress
that we are only refocusing EVSIDS rather than trying to replace it.
Our hybrid solver based on MiniSat only queries NeuroCore once every 100 seconds.

\section{Solver Experiments}\label{sec:experiments}

We evaluate the hybrid solver \emph{neuro-minisat} (described in \S\ref{sec:hybrid}) and the original MiniSat
solver \emph{minisat} on the 400 problems from the main track of
SATCOMP-2018, with the same 5,000 second timeout used in the
competition. For each solver, we solved the 400 problems in 400
different processes in parallel, spread out over 8 identical 64-core
machines, with no other compute-intensive processes running on any of
the machines. In addition, the hybrid solver also had network access
to 5 machines each with 4 GPUs, with the 20 GPUs split evenly and
randomly across the 400 processes. We calculate the running time of a
solver by adding together its process time with the sum of the
wall-clock times of each of the TensorFlow queries it requests on the
GPU servers. We ignore the network transmission times since in
practice one would often use an on-device hardware accelerator.

Note that although we did not train NeuroCore on any (sub)problems
from SATCOMP-2018, we did perform some extremely coarse tuning of
hyperparameters (specifically $\kappa$, which a-priori might reasonably
span 100 orders of magnitude) based on runs of the hybrid solver on
problems from SATCOMP-2018. In hindsight we regret not using
alternate problems for this, but we strongly suspect that we would have
found a similar ballpark by only tuning on problems from other
sources.

\begin{figure}
\centering
\begin{minipage}{2.0in}
  \begin{center}
\includegraphics[width=\textwidth]{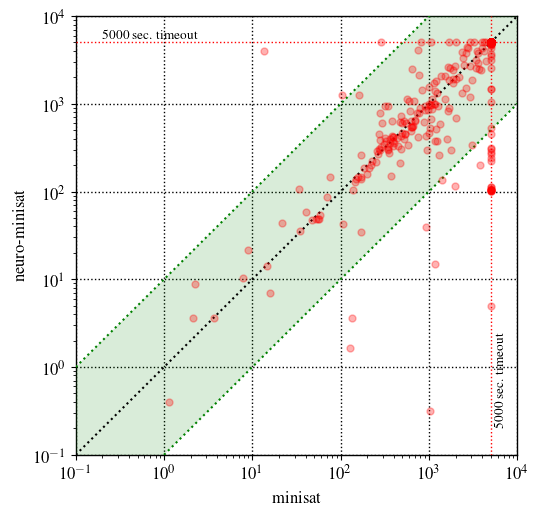}
\caption{Scatter plot comparing NeuroCore-assisted MiniSat (\emph{neuro-minisat}) against (\emph{minisat}).
Several problems are solved within a few minutes by \emph{neuro-minisat}
for which \emph{minisat} times out. The converse scenario is relatively rare.}
\label{fig:scatter}
  \end{center}
\end{minipage}
\quad
\begin{minipage}{2.5in}
 \begin{center}
\includegraphics[width=0.99\textwidth]{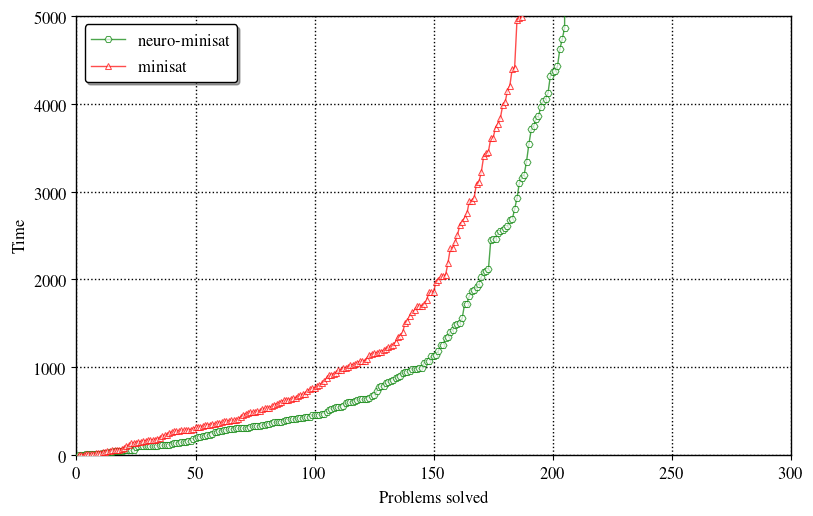}
\caption{Cactus plot comparing NeuroCore-assisted MiniSat (\emph{neuro-minisat}) with the original (\emph{minisat}).
  It shows that \emph{neuro-minisat} takes a substantial within the first few minutes and maintains the lead until the end.} \label{fig:cactus}
 \end{center}
\end{minipage}
\end{figure}

\paragraph{Results.}
The main result, alluded to in \S\ref{sec:intro}, is that \emph{neuro-minisat}
solves 205 problems within the 5,000 second timeout whereas
\emph{minisat} only solves 187. This corresponds to an increase of
10\%. Most of the improvement comes from solving more satisfiable
problems: \emph{neuro-minisat} solve 125 satisfiable problems compared to
\emph{minisat}'s 109, which is a 15\% increase. On the other hand,
\emph{neuro-minisat} only solved 3\% more unsatisfiable problems (80 vs 78).
Figure~\ref{fig:cactus} shows a cactus plot of the two solvers, which
shows that \emph{neuro-minisat} takes a substantial lead within the first
minutes and maintains the lead until the end.  Figure~\ref{fig:scatter} shows a scatter plot of the same
data, which shows there are quite a few problems that \emph{neuro-minisat}
solves within a few minutes that \emph{minisat} times out on. It also
shows that there are very few problems on which \emph{neuro-minisat} is
substantially worse than \emph{minisat}.

\begin{figure}
  \begin{center}
\includegraphics[width=0.5\textwidth]{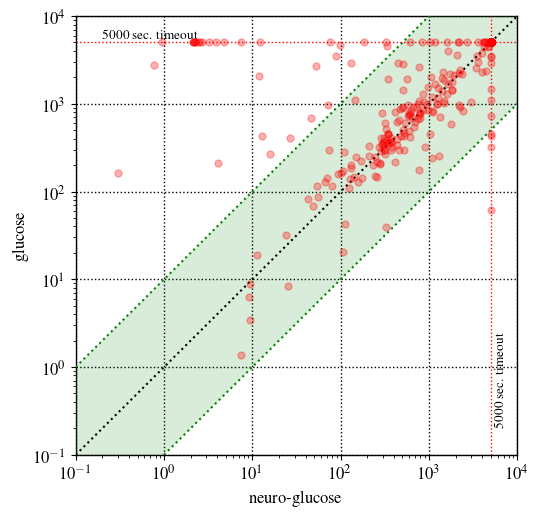}
\caption{Scatter plot comparing NeuroCore-assisted Glucose (\emph{neuro-glucose}) with the original (\emph{glucose}).
It shows that there are quite a few problems that \emph{neuro-glucose}
solves within a few seconds that \emph{glucose} times out on, and there are very few problems on which \emph{neuro-glucose} is
substantially worse than \emph{glucose}.}\label{fig:glucose_scatter}
  \end{center}
\end{figure}

\paragraph{Glucose.}
As a follow-up experiment and sanity-check, we made the same
modifications to Glucose 4.1 and evaluated in the same way on
SATCOMP-2018. To provide further assurance that our findings are
robust, we altered the NeuroCore schedule, changing from fixed pauses (100
seconds) to exponential backoff ($5$ seconds at first with multiplier $\gamma=1.2$).
The results of the experiment are very similar to the results from the
MiniSat experiment described above. The number of problems solved
within the timeout jumps 11\% from 186 to 206. Figure~\ref{fig:glucose_scatter} show the scatter plot comparing \emph{neuro-glucose} to \emph{glucose}.
This comparison is even more favorable to the NeuroCore-assisted solver than Figure~\ref{fig:scatter}, as it shows that there are many problems
\emph{neuro-glucose} solves within seconds that \emph{glucose} times out on. The cactus plot for the Glucose experiment is almost identical to the one in Figure~\ref{fig:cactus} and so is not shown.

\paragraph{Z3.}
Lastly, we made the same modifications to Z3, except we once again
altered the NeuroCore schedule, this time from exponential backoff in
terms of user-time to geometric backoff in terms of the number of
conflicts. Specifically, we first query NeuroCore after
50,000 conflicts, and then each time wait 50,000 more conflicts than
the previous time before querying NeuroCore again. The modified Z3 solves
170 problems within the timeout, up from 161 problems, which is a 6\%
increase.

Note that for the Z3 experiment, to save on computational costs,
we evaluated both solvers simultaneously instead of sequentially.
To ensure fairness, we ordered the task queue by problem rather than by solver.
The lower absolute scores compared to MiniSat and Glucose are partly the
result of the increased contention.

\paragraph{A more favorable regime.}

It is worth remarking that SATCOMP-2018 is an extremely unfavorable regime for machine learning methods.
All problems are arbitrarily out of distribution.
The 2018 benchmarks include problems arising from a dizzyingly diverse set of domains:
proving theorems about bit-vectors,
reversing Cellular Automata, verifying floating-point computations,
finding efficient polynomial multiplication circuits,
mining Bitcoins,
allocating time-slots to students with preferences,
and finding Hamiltonian cycles as part of a puzzle game,
among many others~\cite{satcomp2018}.

In practice, one often wants to solve many problems arising from a common source
over an extended period of time, in which case it could be worth training a neural network specifically
for the problem distribution in question. We approximate this regime by evaluating the same
trained network discussed above on the set of 303 (non-public) hard scheduling problems that
were included in the data generation process along with SATCOMP 2013-2017. Note that although
NeuroCore may have seen the cores of \emph{subproblems} of these problems during training,
most of the problems are so hard that many variables need to be set before Z3 can solve
them in under a minute. Also, at deployment time we are passing the learned clauses to NeuroCore as well,
which may vastly outnumber the original clauses. Thus, although it clearly cannot hurt to train on subproblems
of the test problems, NeuroCore is still being queried on problems that are substantially different
than those it saw during training.

\begin{figure}
  \begin{center}
\includegraphics[width=0.7\textwidth]{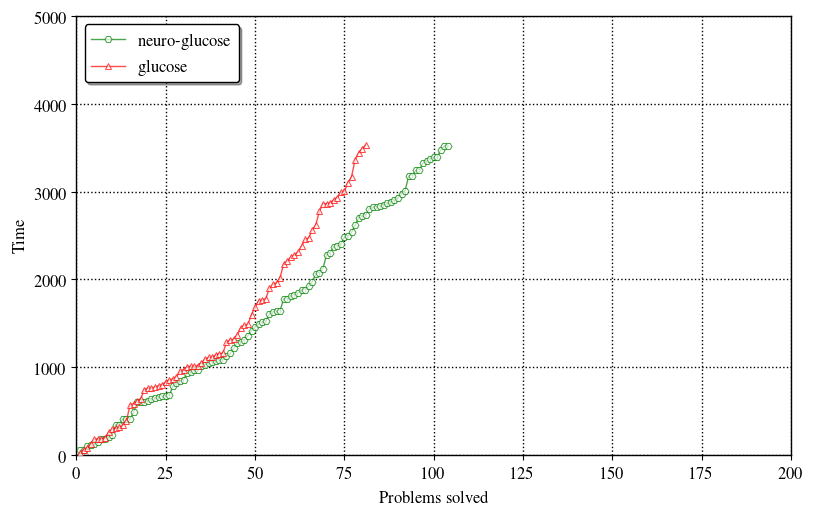}
\caption{Cactus plot comparing NeuroCore-assisted Glucose (\emph{neuro-glucose}) with the original (\emph{glucose})
  on a benchmark of 303 (non-public) challenging scheduling problems, for which some subproblems were included in
  the training set. In contrast to Figure~\ref{fig:cactus}, which showed that on SATCOMP-2018 \emph{neuro-minisat}
got off to an early lead and maintained it throughout, here we see that the solvers are roughly tied
for the first thirty minutes, at which point \emph{neuro-glucose} begins to pull away, and continues
to add to its lead until the one-hour timeout.
}
\label{fig:nc:nfl:cactus}
  \end{center}
\end{figure}

For this experiment, we compared \emph{glucose} to \emph{neuro-glucose} on the 303 scheduling problems,
using a one-hour timeout and the same setting of $\kappa$ as for the SATCOMP-2018 experiment above.
As one might expect, the results are even better than in the SATCOMP regime.
The hybrid \emph{neuro-glucose} solver solves 20\% more problems than \emph{glucose} within the timeout.
Figure~\ref{fig:nc:nfl:cactus} shows a cactus plot comparing the two solvers.
In contrast to Figure~\ref{fig:cactus}, which showed that on SATCOMP-2018 \emph{neuro-minisat}
got off to an early lead and maintained it throughout, here we see that the solvers are roughly tied
for the first thirty minutes, at which point \emph{neuro-glucose} begins to pull away, and continues
to add to its lead until the one-hour timeout.

\paragraph{Ablations.}

A previous version of this paper reported that periodically refocusing with random scores
(as opposed to with the NeuroCore scores) severely degraded the performance of the solver;
however, these findings were the result of an implementation error that caused the activity scores
to be randomly refocused on every single branching decision.\footnote{We thank
  Alvaro Sanchez for bringing this implementation error to our attention.}
We have fixed the error and performed the following revised experiments.
We modified \emph{neuro-glucose} to use uniform random logits in \( [-1, 1] \)
in lieu of the NeuroCore scores, and evaluated the resulting solver (henceforth \emph{rand-glucose}) on
both SATCOMP-2018 and the private scheduling problems. On SATCOMP-2018, \emph{rand-glucose} solved
the same number of problems that \emph{neuro-glucose} did, whereas on the private scheduling problems,
\emph{rand-glucose} solved 2.5\% fewer problems than \emph{glucose} (while \emph{neuro-glucose} solved 20\% more problems than \emph{glucose}).
The impressive performance of \emph{rand-glucose} on SATCOMP-2018 suggests that whatever signal
may have been present in the NeuroCore scores on these wildly out-of-distribution problems
had little impact, and that most (if not all) of the benefit came from the periodic refocusing.
On the other hand, the unimpressive performance of \emph{rand-glucose} on the scheduling problems
suggests that there was indeed substantial signal in the NeuroCore scores in this domain.
Of course, the scheduling ablation does not rule out the possibility that there may be a different
heuristic that could do just as well as NeuroCore without employing a neural network.

\section{Related Work}
Machine learning in automated deduction has been pursued in several guises.
Two established approaches are strategy selection~\cite{DBLP:journals/jair/XuHHL08} and axiom
selection~\cite{DBLP:conf/cade/UrbanSPV08}.

Strategy selection is to our knowledge
mainly applied to setting configuration parameters for SAT,
MIP (Mixed-Integer Programming), TSP (the Traveling Salesman Problem),
and ATP (First-order Automated Theorem Proving) systems, and enjoy
the additional advantage that they can be used in a setting where multiple systems
are combined to approach a virtual best solver%
\footnote{\url{http://ada.liacs.nl/events/sparkle-sat-18/documents/floc-18-sparkle-extended.pdf}}.
ATP systems regularly use strategy selection especially when preparing for
competitions, e.g.~\cite{DBLP:conf/cade/000117}, ever since
Gandalf~\cite{DBLP:journals/jar/Tammet97} won the 1996 ATP competition (known as CASC~\cite{Sut16}) by spending
the first few minutes running a suite of different strategies before
selecting one that appeared to make the most progress.
Strategy selection as composing tactics~\cite{de2013strategy} was pursued
in~\cite{DBLP:conf/nips/BalunovicBV18} to speed up performance
over baseline tactics.

Axiom selection methods help focus search
on a subset of input clauses. The domain-specific Sine~\cite{DBLP:conf/cade/HoderR0V16} method
is a prominent example, and selects axioms that share infrequently-appearing symbols with the goal.
ATP systems rely on clause selection for driving inferences, and a recent
use of machine learning for clause selection~\cite{loos2017deep}
was integrated in the
E theorem prover~\cite{schulz2002brainiac}. In SAT, CDCL solvers select clauses using unit propagation
and conflict analysis, and rely on garbage collection of redundant clauses
to balance available inferences from memory and propagation overhead.
Carefully crafted methods have been introduced to balance different
heuristics within SAT garbage collection, more recently by~\cite{DBLP:conf/sat/Oh15},
combining glue levels with activity scores.
Ostensibly as a reaction to the opacity and complexity of these heuristics,
the CryptoMiniSat solver\footnote{\url{https://github.com/msoos/cryptominisat/}}
has recently integrated machine learning
to eliminate redundant clauses. Similar to our approach, their approach relies on information from
DRAT proofs to relate features
of learned clauses (for example, their glue levels) with their usefulness to a derivation. The CryptoMiniSat version
of DRAT\footnote{\url{https://github.com/msoos/drat-trim/}}
indeed
collects several more features than the original version of DRAT-trim%
\footnote{\url{https://github.com/marijnheule/drat-trim}} that we used in this work.
Their approach then trains a succinct decision tree on this data, that is compiled into a specialized version
of CryptoMiniSat.

Integration of machine learning techniques for branch selection in SAT is to
our knowledge relatively unexplored. The VSIDS heuristic (and its descendants such as EVSIDS) presented a breakthrough
in SAT solving as it amplified branching on variables that would maximize the
conflict-to-branch ratio, thus focusing search within clusters of related clauses.
Several refinements of VSIDS are used in newer SAT solvers, including
CHB (Conflict History Based)~\cite{DBLP:conf/ijcai/LiangKPCG18} and VMTF (Variable Move-To-Front)~\cite{Ryan2004,biere2015evaluating}.
Branch selection heuristics within CDCL solvers are finely tuned for performance
because they are invoked on every decision.

In contrast, look-ahead solvers~\cite{DBLP:series/faia/HeuleM09} afford higher overhead
during a look-ahead phase to identify branch literals, also known as so-called
\emph{cubes} when look-ahead solving is used in the cube-and-conquer paradigm~\cite{heule2018cube}.
Cubes are selected to optimize a carefully crafted metric on clause reduction,
such as weighing variable occurrences inversely by
the sizes of the clauses they appear in~\cite{heule2016solving,heule2018schur,DBLP:series/faia/Kullmann09}.
Cubing is an appealing target for machine learning because it is used in phases
where a global analysis on a problem is feasible.

In MIP solvers, branch-and-bound methods~\cite{lawler1966branch} share many similarities to cube-and-conquer methods in SAT.
Branch operations split the search into separate parts, and are relatively rare, as the main engine in state-of-the-art
MIP solvers remains dual-Simplex, often augmented by interior point methods.
Branching is applied when the linear programming optimization is unable
to find integral values for integer variables. State-of-the-art branching methods in MIP
solvers use heuristics that are related in spirit to look-ahead heuristics:
among a set of candidate branch variables they run a limited (cheap) form of
linear programming and assemble progress metrics for each candidate
variable, and branch on a variable that optimizes a selected metric.
As a common trait, these metrics depend on finely tuned parameters,
and are therefore ripe targets for machine learning
techniques~\cite{DBLP:conf/icml/BalcanDSV18}.

In the backdrop of the related work, the approach we pursue here
is wedged between the fine-grained branching preferences of CDCL solvers
and the single-step branch decisions of look-ahead solvers. NeuroCore
performs a global analysis to predict a good ordering among \emph{all}
unassigned variables, but does this only periodically to allow the
fine-grained built-in heuristics to take over during inferences.
It provides the capability to rehash a search into a different cluster
of clauses where CDCL can perform local tuning.

\section{Discussion}

There is a vast design space for how to train NeuroSAT and how to use
it to guide SAT solvers. This work has focused on only one tiny
point in that design space. We now briefly discuss other approaches we considered.

For our first experiment predicting unsatisfiable cores, we trained NeuroSAT on a synthetic
graph coloring distribution that happened to have tiny
cores. NeuroSAT was able to predict these cores so accurately that we
could get almost arbitrarily big speedups by only giving Z3
the tiny fraction of clauses that NeuroSAT thought most likely to be in the
core (and doubling the number of clauses given as necessary until they
included the core). Unfortunately, it is much harder to learn a
general-purpose core predictor than one on a particular synthetic
distribution for which instances may all have similar cores. Real
problems also rarely have such tiny cores, so even a perfect core
predictor might not be such a silver bullet. However, we do think that core predictions
may nonetheless be useful in guiding clause pruning. Our first efforts here were hampered
by the fact that we were rarely able to fit the majority of conflict clauses in GPU memory
given our relatively large network architecture (\ie $d$=80).
Simply retraining with a smaller $d$ would address this problem, and we plan to pursue this in the future.

We also experimented with training NeuroSAT to imitate the
decisions of the March cubing heuristic. Based on preliminary
experiments in a challenging scheduling domain, we found that NeuroSAT
trained only to imitate March may actually produce better cubes than
March itself, though it remains to be seen if this result holds up to
greater scrutiny. In contrast, using the unsat core predictions to make cubing decisions seemed
to perform consistently worse than the March baseline, though still vastly better than random.
We also tried using NeuroSAT's March predictions to refocus the EVSIDS
scores, and found this to perform worse than its unsat core predictions.
However, we note that the March predictions were much peakier than the unsat core predictions,
and so the inferior performance may have been caused by an inappropriately low temperature parameter $\tau$.

We also briefly experimented with predicting models directly.
Specifically, we used existing solvers to find models of satisfiable problems,
and then trained NeuroSAT to predict the phases of each of the variables individually.
Then, we instrumented MiniSat to choose the phase of each decision variable in proportion
to NeuroSAT's prediction. Note that this approach is extremely simplistic, since
a single problem may have many models; for example, if it suffices to assign only $\epsilon\%$ of the variables
to satisfy all the clauses, then $(1-\epsilon)\%$ of the phases will be arbitrary.
Nonetheless, we still found some preliminary evidence that even this simplistic approach may help in some cases,
though the preliminary results were insufficiently promising for us to pursue further at this stage.
An alternative approach to learning a phase heuristic may be to only predict the phases of variables for which one literal has been proven to be entailed.
It is trivial to generate a huge amount of data for this task, since every learned conflict clause provides datapoints:
each literal in the learned clause is implied from the remaining literals negated.

Lastly, inspired by the success of~\cite{silver2017mastering}, we
experimented with various forms of Monte Carlo tree search and
reinforcement learning, though the only competitive heuristic we
were able to learn \emph{de novo} was a cubing strategy for uniform random
problems.  There are two main challenges for learning variable
branching heuristics by exploration alone: problems may have a huge
number of variables, and it may take substantial time to solve
the (sub)problems in order to get feedback about a given branching
decision. The former challenge can be mitigated by beginning with
imitation learning (e.g. by imitating March). We tried to mitigate the
latter by pretraining a value function based on data collected from
solving a collection of benchmarks, and then using the value function
estimates to make cheap importance sampling estimates of the size of
the search tree under different policies as described
in~\cite{knuth1975estimating}. We found that even in the supervised
context, training the value function was difficult; without taking
$\log$s it was numerically difficult, and with taking $\log$s, we could
get very low loss while ignoring the relatively few hard subproblems
towards the roots that make the most difference.
Ultimately, we think that the use of clausal proof traces for
satisfiability problems offers such great opportunities
for \emph{post facto} analysis and principled credit assignment that
there is simply no need to resort to generic reinforcement learning methods.

We have only scratched the surface of this design space. We hope that
our promising initial results with NeuroCore inspire others to try
leveraging NeuroSAT in other, creative ways.

\section{Acknowledgments}

We thank Percy Liang, David L. Dill, and Marijn J. H. Heule for helpful discussions.


\bibliographystyle{splncs04}
\bibliography{main}

\begin{thebibliography}{10}
\providecommand{\url}[1]{\texttt{#1}}
\providecommand{\urlprefix}{URL }
\providecommand{\doi}[1]{https://doi.org/#1}

\bibitem{abadi2016tensorflow}
Abadi, M., Barham, P., Chen, J., Chen, Z., Davis, A., Dean, J., Devin, M.,
  Ghemawat, S., Irving, G., Isard, M., et~al.: Tensorflow: A system for
  large-scale machine learning. In: 12th {USENIX} Symposium on Operating
  Systems Design and Implementation {OSDI} 16. pp. 265--283 (2016)

\bibitem{DBLP:conf/icml/BalcanDSV18}
Balcan, M., Dick, T., Sandholm, T., Vitercik, E.: Learning to branch. In: Dy,
  J.G., Krause, A. (eds.) Proceedings of the 35th International Conference on
  Machine Learning, {ICML} 2018, Stockholmsm{\"{a}}ssan, Stockholm, Sweden,
  July 10-15, 2018. {JMLR} Workshop and Conference Proceedings, vol.~80, pp.
  353--362. JMLR.org (2018),
  \url{http://proceedings.mlr.press/v80/balcan18a.html}

\bibitem{DBLP:conf/nips/BalunovicBV18}
Balunovic, M., Bielik, P., Vechev, M.T.: Learning to solve {SMT} formulas. In:
  Bengio, S., Wallach, H.M., Larochelle, H., Grauman, K., Cesa{-}Bianchi, N.,
  Garnett, R. (eds.) Advances in Neural Information Processing Systems 31:
  Annual Conference on Neural Information Processing Systems 2018, NeurIPS
  2018, 3-8 December 2018, Montr{\'{e}}al, Canada. pp. 10338--10349 (2018),
  \url{http://papers.nips.cc/paper/8233-learning-to-solve-smt-formulas}

\bibitem{biere2015evaluating}
Biere, A., Fr{\"o}hlich, A.: Evaluating {CDCL} variable scoring schemes. In:
  International Conference on Theory and Applications of Satisfiability
  Testing. pp. 405--422. Springer (2015)

\bibitem{DBLP:series/faia/2009-185}
Biere, A., Heule, M., van Maaren, H., Walsh, T. (eds.): Handbook of
  Satisfiability, Frontiers in Artificial Intelligence and Applications,
  vol.~185. {IOS} Press (2009)

\bibitem{de2013strategy}
De~Moura, L., Passmore, G.O.: The strategy challenge in {SMT} solving. In:
  Automated Reasoning and Mathematics, pp. 15--44. Springer (2013)

\bibitem{devlin2017robustfill}
Devlin, J., Uesato, J., Bhupatiraju, S., Singh, R., Mohamed, A.r., Kohli, P.:
  {Robustfill: Neural program learning under noisy I/O}. In: Proceedings of the
  34th International Conference on Machine Learning-Volume 70. pp. 990--998.
  JMLR. org (2017)

\bibitem{DBLP:series/faia/HeuleM09}
Heule, M., van Maaren, H.: Look-ahead based {SAT} solvers. In: Biere et~al.
  \cite{DBLP:series/faia/2009-185}, pp. 155--184.
  \doi{10.3233/978-1-58603-929-5-155}

\bibitem{heule2018schur}
Heule, M.J.: Schur number five. In: Thirty-Second AAAI Conference on Artificial
  Intelligence (2018)

\bibitem{heule2018cube}
Heule, M.J., Kullmann, O., Biere, A.: Cube-and-conquer for satisfiability. In:
  Handbook of Parallel Constraint Reasoning, pp. 31--59. Springer (2018)

\bibitem{heule2016solving}
Heule, M.J., Kullmann, O., Marek, V.W.: Solving and verifying the boolean
  pythagorean triples problem via cube-and-conquer. In: International
  Conference on Theory and Applications of Satisfiability Testing. pp.
  228--245. Springer (2016)

\bibitem{satcomp2018}
Proceedings of sat competition 2018; solver and benchmark descriptions  (2018),
  \url{http://hdl.handle.net/10138/237063}

\bibitem{hinton2012deep}
Hinton, G., Deng, L., Yu, D., Dahl, G., Mohamed, A.r., Jaitly, N., Senior, A.,
  Vanhoucke, V., Nguyen, P., Kingsbury, B., et~al.: Deep neural networks for
  acoustic modeling in speech recognition. IEEE Signal processing magazine
  \textbf{29} (2012)

\bibitem{hochreiter1997lstm}
Hochreiter, S., Schmidhuber, J.: Long short-term memory. Neural Computation
  \textbf{9}(8),  1735--1780 (1997)

\bibitem{DBLP:conf/cade/HoderR0V16}
Hoder, K., Reger, G., Suda, M., Voronkov, A.: Selecting the selection. In:
  Olivetti, N., Tiwari, A. (eds.) Automated Reasoning - 8th International Joint
  Conference, {IJCAR} 2016, Coimbra, Portugal, June 27 - July 2, 2016,
  Proceedings. Lecture Notes in Computer Science, vol.~9706, pp. 313--329.
  Springer (2016). \doi{10.1007/978-3-319-40229-1\_22}

\bibitem{huang2018gamepad}
Huang, D., Dhariwal, P., Song, D., Sutskever, I.: Gamepad: A learning
  environment for theorem proving. arXiv preprint arXiv:1806.00608  (2018)

\bibitem{irving2016deepmath}
Irving, G., Szegedy, C., Alemi, A.A., Een, N., Chollet, F., Urban, J.:
  Deepmath-deep sequence models for premise selection. In: Advances in Neural
  Information Processing Systems. pp. 2235--2243 (2016)

\bibitem{kaliszyk2017holstep}
Kaliszyk, C., Chollet, F., Szegedy, C.: Holstep: A machine learning dataset for
  higher-order logic theorem proving. arXiv preprint arXiv:1703.00426  (2017)

\bibitem{kingma2014adam}
Kingma, D.P., Ba, J.: Adam: A method for stochastic optimization. arXiv
  preprint arXiv:1412.6980  (2014)

\bibitem{knuth1975estimating}
Knuth, D.E.: Estimating the efficiency of backtrack programs. Mathematics of
  computation  \textbf{29}(129),  122--136 (1975)

\bibitem{knuth2015sat}
Knuth, D.E.: {The Art of Computer Programming, Volume 4, Fascicle 6:
  Satisfiability} (2015)

\bibitem{krizhevsky2012imagenet}
Krizhevsky, A., Sutskever, I., Hinton, G.E.: Imagenet classification with deep
  convolutional neural networks. In: Advances in neural information processing
  systems. pp. 1097--1105 (2012)

\bibitem{kullback1951information}
Kullback, S., Leibler, R.A.: On information and sufficiency. The annals of
  mathematical statistics  \textbf{22}(1),  79--86 (1951)

\bibitem{DBLP:series/faia/Kullmann09}
Kullmann, O.: Fundaments of branching heuristics. In: Biere et~al.
  \cite{DBLP:series/faia/2009-185}, pp. 205--244.
  \doi{10.3233/978-1-58603-929-5-205}

\bibitem{lawler1966branch}
Lawler, E.L., Wood, D.E.: Branch-and-bound methods: A survey. Operations
  research  \textbf{14}(4),  699--719 (1966)

\bibitem{DBLP:conf/ijcai/LiangKPCG18}
Liang, J., K., H.G.V., Poupart, P., Czarnecki, K., Ganesh, V.: An empirical
  study of branching heuristics through the lens of global learning rate. In:
  Lang, J. (ed.) Proceedings of the Twenty-Seventh International Joint
  Conference on Artificial Intelligence, {IJCAI} 2018, July 13-19, 2018,
  Stockholm, Sweden. pp. 5319--5323. ijcai.org (2018).
  \doi{10.24963/ijcai.2018/745}, \url{http://www.ijcai.org/proceedings/2018/}

\bibitem{loos2017deep}
Loos, S.M., Irving, G., Szegedy, C., Kaliszyk, C.: Deep network guided proof
  search. In: Eiter, T., Sands, D. (eds.) LPAR-21, 21st International
  Conference on Logic for Programming, Artificial Intelligence and Reasoning,
  Maun, Botswana, May 7-12, 2017. EPiC Series in Computing, vol.~46, pp.
  85--105. EasyChair (2017),
  \url{http://www.easychair.org/publications/paper/340345}

\bibitem{marques1999grasp}
Marques-Silva, J.P., Sakallah, K.A.: Grasp: A search algorithm for
  propositional satisfiability. IEEE Transactions on Computers  \textbf{48}(5),
   506--521 (1999)

\bibitem{Mijnders2010SymbiosisOS}
Mijnders, S., de~Wilde, B., Heule, M.: Symbiosis of search and heuristics for
  random 3-sat. CoRR  \textbf{abs/1402.4455} (2010)

\bibitem{moritz2018ray}
Moritz, P., Nishihara, R., Wang, S., Tumanov, A., Liaw, R., Liang, E., Elibol,
  M., Yang, Z., Paul, W., Jordan, M.I., et~al.: Ray: A distributed framework
  for emerging $\{$AI$\}$ applications. In: 13th {USENIX} Symposium on
  Operating Systems Design and Implementation {OSDI} 18). pp. 561--577 (2018)

\bibitem{moskewicz2001chaff}
Moskewicz, M.W., Madigan, C.F., Zhao, Y., Zhang, L., Malik, S.: Chaff:
  Engineering an efficient sat solver. In: Proceedings of the 38th annual
  Design Automation Conference. pp. 530--535. ACM (2001)

\bibitem{DBLP:conf/sat/Oh15}
Oh, C.: Between {SAT} and {UNSAT:} the fundamental difference in {CDCL} {SAT}.
  In: Heule, M., Weaver, S. (eds.) Theory and Applications of Satisfiability
  Testing - {SAT} 2015 - 18th International Conference, Austin, TX, USA,
  September 24-27, 2015, Proceedings. Lecture Notes in Computer Science,
  vol.~9340, pp. 307--323. Springer (2015). \doi{10.1007/978-3-319-24318-4\_23}

\bibitem{parisotto2016neuro}
Parisotto, E., Mohamed, A.r., Singh, R., Li, L., Zhou, D., Kohli, P.:
  Neuro-symbolic program synthesis. arXiv preprint arXiv:1611.01855  (2016)

\bibitem{Ryan2004}
Ryan, L.: Efficient algorithms for clause-learning {SAT} solvers (2004),
  masters thesis

\bibitem{schulz2002brainiac}
Schulz, S.: E--a brainiac theorem prover. Ai Communications  \textbf{15}(2, 3),
   111--126 (2002)

\bibitem{DBLP:conf/cade/000117}
Schulz, S.: We know (nearly) nothing! but can we learn? In: Reger, G., Traytel,
  D. (eds.) {ARCADE} 2017, 1st International Workshop on Automated Reasoning:
  Challenges, Applications, Directions, Exemplary Achievements, Gothenburg,
  Sweden, 6th August 2017. EPiC Series in Computing, vol.~51, pp. 29--32.
  EasyChair (2017), \url{http://www.easychair.org/publications/paper/6kgF}

\bibitem{selsam2018learning}
Selsam, D., Lamm, M., B\"{u}nz, B., Liang, P., de~Moura, L., Dill, D.L.:
  Learning a {SAT} solver from single-bit supervision. In: International
  Conference on Learning Representations (2019),
  \url{https://openreview.net/forum?id=HJMC\_iA5tm}

\bibitem{silver2017mastering}
Silver, D., Schrittwieser, J., Simonyan, K., Antonoglou, I., Huang, A., Guez,
  A., Hubert, T., Baker, L., Lai, M., Bolton, A., et~al.: Mastering the game of
  go without human knowledge. Nature  \textbf{550}(7676),  354--359 (2017)

\bibitem{Sut16}
Sutcliffe, G.: {The CADE ATP System Competition - CASC}. AI Magazine
  \textbf{37}(2),  99--101 (2016)

\bibitem{DBLP:journals/jar/Tammet97}
Tammet, T.: Gandalf. J. Autom. Reasoning  \textbf{18}(2),  199--204 (1997).
  \doi{10.1023/A:1005887414560}

\bibitem{DBLP:conf/cade/UrbanSPV08}
Urban, J., Sutcliffe, G., Pudl{\'{a}}k, P., Vyskocil, J.: Malarea {SG1-}
  machine learner for automated reasoning with semantic guidance. In: Armando,
  A., Baumgartner, P., Dowek, G. (eds.) Automated Reasoning, 4th International
  Joint Conference, {IJCAR} 2008, Sydney, Australia, August 12-15, 2008,
  Proceedings. Lecture Notes in Computer Science, vol.~5195, pp. 441--456.
  Springer (2008). \doi{10.1007/978-3-540-71070-7\_37}

\bibitem{wang2017premise}
Wang, M., Tang, Y., Wang, J., Deng, J.: Premise selection for theorem proving
  by deep graph embedding. In: Advances in Neural Information Processing
  Systems. pp. 2786--2796 (2017)

\bibitem{wetzler2014drat}
Wetzler, N., Heule, M.J., Hunt, W.A.: Drat-trim: Efficient checking and
  trimming using expressive clausal proofs. In: International Conference on
  Theory and Applications of Satisfiability Testing. pp. 422--429. Springer
  (2014)

\bibitem{whalen2016holophrasm}
Whalen, D.: Holophrasm: a neural automated theorem prover for higher-order
  logic. arXiv preprint arXiv:1608.02644  (2016)

\bibitem{wu2016google}
Wu, Y., Schuster, M., Chen, Z., Le, Q.V., Norouzi, M., Macherey, W., Krikun,
  M., Cao, Y., Gao, Q., Macherey, K., et~al.: Google's neural machine
  translation system: Bridging the gap between human and machine translation.
  arXiv preprint arXiv:1609.08144  (2016)

\bibitem{DBLP:journals/jair/XuHHL08}
Xu, L., Hutter, F., Hoos, H.H., Leyton{-}Brown, K.: Satzilla: Portfolio-based
  algorithm selection for {SAT}. J. Artif. Intell. Res.  \textbf{32},  565--606
  (2008). \doi{10.1613/jair.2490}

\end{thebibliography}
\end{document}